# Spelling Error Trends and Patterns in Sindhi

[1] Zeeshan Bhatti,[2] Imdad Ali Ismaili, [3] Asad Ali Shaikh, [4] Waseem Javaid
[1,2,3,4] Institute of Information and Communication Technology, University of Sindh, Jamshoro
Email :{ [1] zeeshanbhatti22@hotmail.com, [2] iai_a@yahoo.com,[3] dr.asadshaikh@yahoo.com,[4] waseem.javaid@usindh.edu.pk}

## ABSTRACT

Statistical error Correction technique is the most accurate and widely used approach today, but for a language like Sindhi which is a low resourced language the trained corpora's are not available, so the statistical techniques are not possible at all. Instead a useful alternative would be to exploit various spelling error trends in Sindhi by using a Rule based approach. For designing such technique an essential prerequisite would be to study the various error patterns in a language. This paper presents various studies of spelling error trends and their types in Sindhi Language. The research shows that the error trends common to all languages are also encountered in Sindhi but their do exist some error patters that are catered specifically to a Sindhi language.

## 1. INTRODUCTION

Spell Checker have been developed for many world languages. The research in the field of detecting errors in typed words started in 1960 [2]. The investigations have been made that the first spell checkers were only 'Verifiers'. They were not performing the task of 'Correctors', in that they offered no suggestions for incorrect spelling word. The investigation showed that for logical or phonetic errors those verifiers were not helpful as they only helped in typo mistakes. Thus the early developers faced immense challenges and difficulties in offering useful suggestions for misspelled words. Thus for the understanding and implementation of an advance spell checking algorithm various techniques and algorithms have been developed for various languages of the world.

Sindhi language spoken in Pakistan and some parts of India is one of the oldest language of the world, using Persio-Arabic and Devnagri script. Due to its script and ligature complexity, it's very essential to understand the ligature structure and analyze the various error patterns or trends in Sindhi language before any attempt on developing a spell checker could be made.

## 2. THE WORD "ERROR"

There are two distinct categories in which the word 'Error' can be placed, first being the "Non-Word Error" and the second category is the "Real-Word Error". Consider a candidate string- which is collection of characters and usually separated by hard space or punctuation marks, if this string or word has a meaning then it would be called as a candidate string otherwise this string is referred to as a 'Non-Word Error'[2]. Real word errors are considered to be such type of errors that occur when the typist or writer types a valid word which is spelled correctly but it is not the actually word intended to be written with respect to context. Therefore the sentence would be considered incorrect syntactically or semantically [3].

## 3. CLASSIFICATIONS OF ERRORS

In the literature, there are several theories on various classifications of errors and each theory does not exclude another. Several of these error classification theories have been discussed by Kukich [4]. In the next subsection (3.1), I review the various different classification theories from several different studies. In section 3.2, the classifications of errors from these studies are discussed.

### 3.1 Typing Errors And Spelling Errors

The basic type of classification of errors is between typing errors (commonly known as 'typos') and spelling errors. Typing errors occurs when the typist or the author knows the actual and correct spelling of the word but mistakenly or by slip of finger presses an invalid key. It has been observed that since the writer is aware of the actually spelling of the words so the possibilities of multiple mistakes within the word are very low and the resulting misspelled word will usually resemble orthographically to the intended word. (e.g. صپريم ڪورٽ {cupreme court} for سپريم ڪورٽ {Supreme Court (noun): supreme kourt}).

Whereas the term spelling errors refers to the errors that occur due to the ignorance of the author or typist regardless of the fact that the actual spelling of the intended word is known or not. Usually the terms spelling errors and typing errors are considered to be same and referred to simply as a spelling errors, this sometimes also creates confusion among the users. It has been observed that there are three general possibilities of spelling errors:

i. Phonetic error are errors due to the similar pronounced characters in a language where the incorrect or misspelled word is phonetically similar to the intended word (e.g. rof ماڻهوماڊ حو {human: mardhon} or ناتصڪاپ for پاڪستان {Pakistan (noun): Pakistan}) and so the phonetically incorrect word is sometimes also referred to as a homophone of the actual intended word.
ii. The other reason for the spelling mistake is that the incorrect or misspelled word is semantically similar to the actual intended word (e.g. حيحص for يحص {correct: sahi} or وتحالص for وحالص {guider : salahoo} or يحالص ).
iii. The last reason behind the spelling errors is that the writer or the typist is ignorant or unaware of the basic grammatical rules of a language while typing the text. According to researchers and linguistics the grammatical errors do not fall under the category of spelling errors but according to them such type of errors should be considered under a separate category and so usually most of the general spelling error detection tools and software do not consider grammatical errors and ignore them.

## 3.2 Single Error Words And The Multiple Error Words

From the literature review of kukich [4] the second category of errors discussed is the distinction between the words that contain multiple errors against just a single character error, hence the terms single error words and multiple error words. Kukick in his studies describes that single error mistakes usually occur due to one of the following four reasons [4]:

i. A single character inserted at invalid position in a string.
ii. A single character deleted form an intended word.
iii. A character substituted by another incorrect or undesired character.
iv. Character transposition among two letters in a word.

These are also sometimes collectively referred to as 'single transformations' and are also known as 'Damerau transformations' [5]. Words that include more than one of the above mentioned error types are simply known as multiple error words.

## 3.3 Errors In Long Words And The Errors In Short Words

The next category of error classification is the distinction between the errors in short words and the errors that are found in long words or a valid string that consists of large number of characters. It has been analyzed from the studies of Pollock and Zamora [6] that in short or small words - with three or four characters, there is an extensive variance in the amount of errors that occur that constitutes 9.2% of all errors. According to Yannakoudakis and Fawthrop [7] the frequency of errors in short words is 1.5%, however Kukich [8] in his studies disagrees with previous two analysis and emphasizes that the frequency of errors is 63% in small words of two, three and four characters.

## 3.4 First Character Errors And The Nth Character Errors

The next type of error classification discussed in the literature is the distinction between the error that occurs in the first or initial character of the word and the error that occurs in rest of the word or in all other characters of the word except the first. The first or the initial character of the intended word can be deleted ( فاظت ـ rof فاظت ف {protection: hifazat}, substituted (تظافھ for تظافح {protection: hifazat}), transposed with the second character (!فاظت ـprotection: hifazat}) or some other character or letter can be inserted before the first or initial character (حج rof فاظت ف:noitcetorp} فاظت ـ hifazat}). An nth error of a character in a word occurs when there is an error in certain characters of words other than the first or initial character. In nth error type first character must be correct.

## 3.5 Errors Within One Word And The Word Boundary Errors

In this type of error classifications, the distinction between errors within a Sindhi word and errors at the boundary of multiple words, are discussed. The errors due to the Word boundary are such type of errors that occur due to the result of incorrect or misplaced spacing of words. There are generally two types or reasons behind the word boundary error:

i. Incorrect splits, when the space character is inserted at an invalid position of a word hence splitting a single word into two, e.g. ڄ امشورو or چام شورو for ڄامشورو {Jamshoro (noun): jamshoro}
ii. Run-ons, when the space character is omitted or not inserted between two separate words hence resulting in the joining of two words as a single word e.g. يونيورسٽيجو for يونيورسٽي جو {University(noun) : uoniversity}.

15% of all the errors discovered by Kukich [9] in his study were actually word boundary errors and similarly Mitton [10] in his research also found that 13% of the errors generated were word boundary errors.

## 3.6 The Real-Word Errors And Non-Word Errors

In the sixth and the last classification of errors a difference is established between the real-word errors and non-word errors. The Non-word errors are such type of errors that are completely invalid and incorrect words not to be found in any dictionary or in any other corpus, (e.g. eht saerehW .({tazafih : noitcetorp} فاظت ـ rof فاظت ھ Real-word errors actually yields a string of characters which is a valid and correct word itself found within a dictionary with some meaning (e.g. حيح for w rekcehc lleps eht ecneh ({ihas : tcerroc}صحيill not be able to identify such type of errors.

Peterson [11] is amongst the few who had investigated and produced several theories on the frequency of real-word errors. Peterson [11] in his research work studied and analyzed the relationship between the lexicon size and real-word error rate, considering only single error misspellings [5].

## 4. SPELLING ERROR TRENDS IN SINDHI

The common technique of developing the spell correcting algorithm is by initially studying and analyzing he spelling error trends (also known as error patterns) in a language. According to Damerau [5] and Peterson [11] the spelling errors in English language are divided into two main categories, typographic errors and cognitive errors. Through research and analysis I have ascertained that in Sindhi Language Four types of Errors usually occur.

- Typographic Errors
- Cognitive or Phonetic Errors
- Visual Errors
- Space Related Errors

Single-Error is therefore refereed to any type of errors produced from above editing operations [4]. The proclamation made by the Damerau's was later confirmed by a number of researchers and scholar including Peterson [11]. Table 1 shows the results of a study by Peterson [11]. The source of data and information were taken actually from the Webster's Dictionary and Government Printing Office (GPO) documents, which were in-fact retyped by college students [1].

**Table 1:** Statistics of the Four Basic Types of Errors (for English language)

| Error Type | GPO | Web7 |
|---|---|---|
| Transposition | 4 (2.6%) | 47 (13.1%) |
| Insertion | 29 (18.7%) | 73 (20.3%) |
| Deletion | 49 (31.6%) | 124 (34.4%) |
| Substitution | 62 (40.0%) | 97 (26.9%) |
| Total | 144 (92.9%) | 4.7%) |

## 5. TYPOGRAPHIC ERRORS

Typographic Errors occur when a word is mistyped by mistake usually because a finger was placed in the wrong position, whereas the actual and correct spelling of the intended word is known. These type of errors are usually and most often related and dependent on the keyboard, along with the typing knowledge and accuracy of the typist and so such type of errors do not fall under the criteria of any linguistics rule [12][13]. The error patterns here are somewhat dependent on the language and keyboard layout. Damerau [5] in his study emphasized that approximately 80% of the typographic errors fall into one of the following four categories

- **Letter Insertion Error:** While writing a word if a character is pressed twice or some other character is pressed along with the original character, the undesired character is considered to be inserted e.g. ro { tazafih : noitcetorp} فاظت ح rof فاظت سنهذيجح for سنڌي {Sindhi(noun) : sindhi}.

- **Letter Deletion Error:** While writing a word if any or more character is missing then it is called to be deleted and the error is considered to be a deletion error or Omission error e.g. تظاف for ro { tazafih : noitcetorp}فاظت پاڪتانح for پاڪستان {Pakistan (noun): Pakistan}.

- **Letter Substitution Error:** When the desired character in a word got replaced by some other character that is the actual character is substituted. e.g. پارلعامينٽ for پارلياميٽ.{parliament (noun) : parliament}

- **Letter Transposition Error:** As the name suggest, transposition errors occur when characters have "transposed" - that is, they have switched places. Transposition errors are almost always human in origin. The most common way for characters to be transposed is when a user is touch typing at a speed that makes them input one character, before the other. This may be caused by their brain being one step ahead of their body. e.g. تظاحف for تظافح {protection : hifazat } or قئاداعظم for قائداعظم.{Quaid-e-Azam (noun) : quaideazam}

**Table 2:** Various Typographic Error Types

| Error Type | Wrong Word | English Transliteration | Correct Word |
|---|---|---|---|
| Letter Insertion | سنهڌي | Sindhi | سنڌي |
| Letter Omission | پاڪتان | Pakistan | پاڪستان |
| Letter Substitution | پارلعامينٽ | Parliament | پارليامينٽ |
| Transposition of two adjacent letters | قئاداعظم | Quaideazam | قائداعظم |

## 6. COGNITIVE OR PHONETIC ERRORS

Cognitive Errors or Phonetic errors usually occur due to similarly pronounced characters or homophone characters in Sindhi sometimes also referred to as Hearing Errors. Generally the writer is unaware or does not know the actual spelling of the word and the writer tries to write it with probable spelling guessing it with similar pronunciation of the letters that sounds somewhat similar to the actual word. The similar sounding or homophone letters are often substituted in this type of error (e.g. 'ھ' {hah} for 'ح' {hey} or 'ت' {tey} for 'ط' {toye}. e.g. "تاريڪ" {darkness: tareeik} replaced by "طاريڪ". Table 3 shows the list of all the Sindhi letters groped according to their similar sound of pronunciation

**Table 3:** Similarly pronounced letters in Sindhi

| Sound code | Similar Sound Character Group | | | | | |
|---|---|---|---|---|---|---|
| 1 | | ئ | ي | ء | آ | ا |
| 2 | | | | | | ٻ |
| 3 | | | | ڀ | ب | ٻ |
| 4 | | | | | ط | ت |
| 5 | | | | | | ٿ |
| 6 | | | | | ٽ | ٺ |
| 7 | | | | ص | س | ث |
| 8 | | | | چ | ڇ | ج |
| 9 | | | | | ڄ | ڃ |
| 10 | | | | ه | ھ | ح |
| 11 | | | | | ک | خ |
| 12 | | | ڍ | ڏ | ڌ | د |
| 13 | | | ظ | ض | ز | ذ |
| 14 | | | | ر | ڻ | ڙ |
| 15 | | | | | | ش |
| 16 | | | | | ڦ | ف |
| 17 | | | | | ڪ | ق |
| 18 | | | | | غ | گ |
| 19 | | | | | | ل |
| 20 | | | | | | م |
| 21 | | | | | | ن |
| 22 | | | | | ڱ | ڳ |

Some of the error that usually occurs due to the similar sound of Sindhi letters is shown in the table 4 below. Here the writer tries to guess the correct word by inserting a letter which sounds appropriate for the word. The correct word is shown in the left column and their misspelled word is shown in the opposite column.

**Table 4:** Various Examples of cognitive errors

| Wrong Word | English Transliteration | Correct word | |
|---|---|---|---|
| مفروزو or مفروذو | Mafroozo | مفروضو | 1 |
| حڪومط or هڪومت | Hukomat | حڪومت | 2 |
| قرارداذ or ڪرارداد | Qrrardaad | قرارداد | 3 |
| موجودگھي or موجھودگي | Moujeedgy | موجودگي | 4 |

## 7. VISUAL ERRORS

Visual errors are the errors that occur due to the similar shapes of various Sindhi characters. They could also be termed as Reading Errors or similar shape errors. According to Tahira Naseem "Similar shape based errors are not cognitive in nature" [1]. There could be several reasons for this type of errors. First, the professional typists, when typing form a hand written draft of a document will usually type in the text exactly as it looks in his first glaze without giving much consideration to its meaning or word structure and thus characters with similar visual representation are confused with each other rendering in spelling mistakes. Second, when a novice typist or writer of Sindhi language writes a document he may know the correct spelling of the word but could confuse the letters due to shape similarity for example ج {jey} could be mistyped for چ {chey} or ٿ {for {sey ٽ .{Tey}. Third reason could be that the writer is unaware of the correct spelling of the word and tries to guess the word based on the similar shape of characters. Table 5 shows few examples of such type of Visual errors in Sindhi language.

**Table 5:** Types of Error that occur

| Wrong Word | English Transliteration | Correct Word |
|---|---|---|
| دنڏو or دندو | Dhandho | ڌندو |
| ڃڄ or جڄ | Jyunj | ڄڃ |
| ٻڪزي or بڪري | Bakri | ٻڪري |
| احتيازات or اخثيارات | Akhtyrait | اختيارات |

The various shape groups for Sindhi are given in Table 6. Some of the characters in Sindhi language do not share their shapes with any other character and so are so placed alone in the table, there are ten such characters (i.e.: ا , جھ ,ق,ڪ,گھ,ل,م,و,ھ,ء,ي) . Other then these all other characters in Sindhi alphabet have similar shape characters as shown in Table 6 below.

**Table 6:** Basic Shape groups in Sindhi

| S.No: | Base Shape | Shape Group | No: of Elements |
|---|---|---|---|
| 1 | ا | ا | 1 |
| 2 | ٮ | ٿ , ٽ , ٺ, ث<br>ب , ٻ , ڀ , پ, ت | 9 |
| 3 | ح | , ڃ , ڄ , ڇ , چ , ح , خ<br>ج | 7 |
| 4 | حه | جھ | 1 |
| 5 | د | د, ڌ, ڏ, ڊ , ڍ , ذ | 6 |
| 6 | ر | ر , ڙ ,ز | 3 |
| 7 | س | س , ش | 2 |
| 8 | ص | ض<br>ص , | 2 |
| 9 | ط | ط , ظ | 2 |
| 10 | ع | ع , غ | 2 |
| 11 | ڡ | ف , ڦ | 2 |
| 12 | ٯ | ق | 1 |
| 13 | ڪ | ڪ | 1 |
| 14 | ک | ک , گ , ڳ ,ڱ | 4 |
| 15 | گھ | گھ | 1 |
| 16 | ل | ل | 1 |
| 17 | م | م | 1 |
| 18 | ں | ن , ڻ | 2 |
| 19 | و | و | 1 |
| 20 | ه | ھ | 1 |
| 21 | ء | ء | 1 |
| 22 | ى | ي | 1 |
| Total | | | 52 |

## 8. SPACE RELATED ERRORS

Space is one of the most commonly and habitually used character in document drafting and composition. Due to its immense use the involuntary insertion occurs during typing. In Sindhi the space insertion is not as common as space omission which is much more frequent. One reason behind this could be that we usually want to type quickly and want to minimize the efforts hence omitting the space character (or any other character ) intentionally out of ignorance but at the same time we insert a space character either mistakenly or of necessity. This second reason of omitting a space is found to be most common specially when we intent to write complex and long words as Sindhi language has several words that use spaces between two words to make the sentence clear while reading. For example consider the word "لعل شهباز"{Lal Shabaz (noun): lal shahbaaz}, if for instance the space character is omitted or deleted from the middle of the words then the word will become لعلشهباز {LalShabaz(noun): lalshaabaz} which is incorrect and misspelled. Table 7 shows some of the space related errors in Sindhi Language.

**Table 7:** Shows the various space related errors

| Error | Wrong word | English Transliteration | Correct word |
|---|---|---|---|
| Insertion | پا ڙي | Pardein | پاڙي |
| Omission | لعلشهباز | Lal Shabaz | لعل شهباز |
| Transposition | زن دگي | Zindegi | زندگي |

## 9. CONCLUSION

From the discussion and studies given in this paper it can be summarized that in Sindhi, few unique script specific errors trends exists which cannot be found in a language like English. Amongst the various error trends discussed for Sindhi language the one found to be most frequent is substitution errors caused due to the shape similarity of the letters in Sindhi alphabet and also due to the similar pronunciation of various letters. The other type of error found in Sindhi language are the omission or deletion of space character at the word boundaries. From the studied presented for Sindhi it can be assumed that these results and error trends and patters will also apply to other languages that are similar to Persio-Arabic script.